\documentclass[10pt,onecolumn]{article}
\usepackage[total={6.5in, 9.2in}]{geometry}
\usepackage{times}
\usepackage{bbold}
\usepackage{epsfig}
\usepackage{graphicx}
\usepackage{amssymb}
\usepackage{subfiles}
\usepackage{pifont}
\newcommand{\cmark}{\ding{51}}
\newcommand{\xmark}{\ding{55}}
\usepackage{booktabs}
\usepackage{multirow}
\usepackage{graphicx}
\usepackage[normalem]{ulem}
\usepackage{amsmath}
\usepackage{color,soul}
\usepackage{enumitem}
\usepackage[usenames,dvipsnames]{xcolor}
\sethlcolor{pink}
\newcommand{\etal}{\textit{et al.}}
\DeclareMathOperator*{\argmax}{argmax}   
\usepackage[pagebackref=true,breaklinks=true,colorlinks,bookmarks=false]{hyperref}

\begin{document}

\title{\bf Mining bias-target Alignment from Voronoi Cells}

\author{Rémi Nahon
\and
Van-Tam Nguyen
\and
Enzo Tartaglione
}
\date{%
    LTCI, Télécom Paris, Institut Polytechnique de Paris\\%
    \emph{remi.nahon@telecom-paris.fr}\\[2ex]%
}
\maketitle

\begin{abstract}
Despite significant research efforts, deep neural networks are still vulnerable to biases: this raises concerns about their fairness and limits their generalization. In this paper, we propose a bias-agnostic approach to mitigate the impact of bias in deep neural networks. Unlike traditional debiasing approaches, we rely on a metric to quantify ``bias alignment/misalignment'' on target classes, and use this information to discourage the propagation of bias-target alignment information through the network. We conduct experiments on several commonly used datasets for debiasing and compare our method to supervised and bias-specific approaches. Our results indicate that the proposed method achieves comparable performance to state-of-the-art supervised approaches, although it is bias-agnostic, even in presence of multiple biases in the same sample.
\end{abstract}

\section{Introduction}
Deep Neural Networks (DNNs) are known today for their high performance and resilience in many areas of computer vision, such as image classification, semantic segmentation, and object detection, used in areas ranging from self-driving vehicles to face recognition or surgical guidance. However, it is well known that their tendency to rely heavily on any type of correlation present in the training data exposes them to potential pitfalls~\cite{geirhos_texture,blindeye,ood_detection}: some ``spurious correlations''  may be mistakenly learned by the DNN. These can take over the role of \textit{biases}~\cite{bias_in_ds}.

\noindent Learned biases may decrease the generalization of the DNN~\cite{geirhos_texture,blindeye,LNL, LfF, Rebias,rubi}. For example, if a DNN has learned to distinguish airplanes flying in the sky from boats sailing in the ocean, the model will likely use the background as a base for its classification: detecting it instead of learning the vehicle shape is a much simpler task. However, the model does not generalize to scenarios such as a landing seaplane. Differently from domain adaptation~\cite{partial_da,domain_adaptation, domain_invariance}, where the objective is to learn general features compensating the domain shift, or to adapt the extracted features to different domains, the goal of debiasing is to discourage the learning of spurious correlations.

\begin{figure}[t]
\begin{center}
   \includegraphics[width=.8\linewidth]{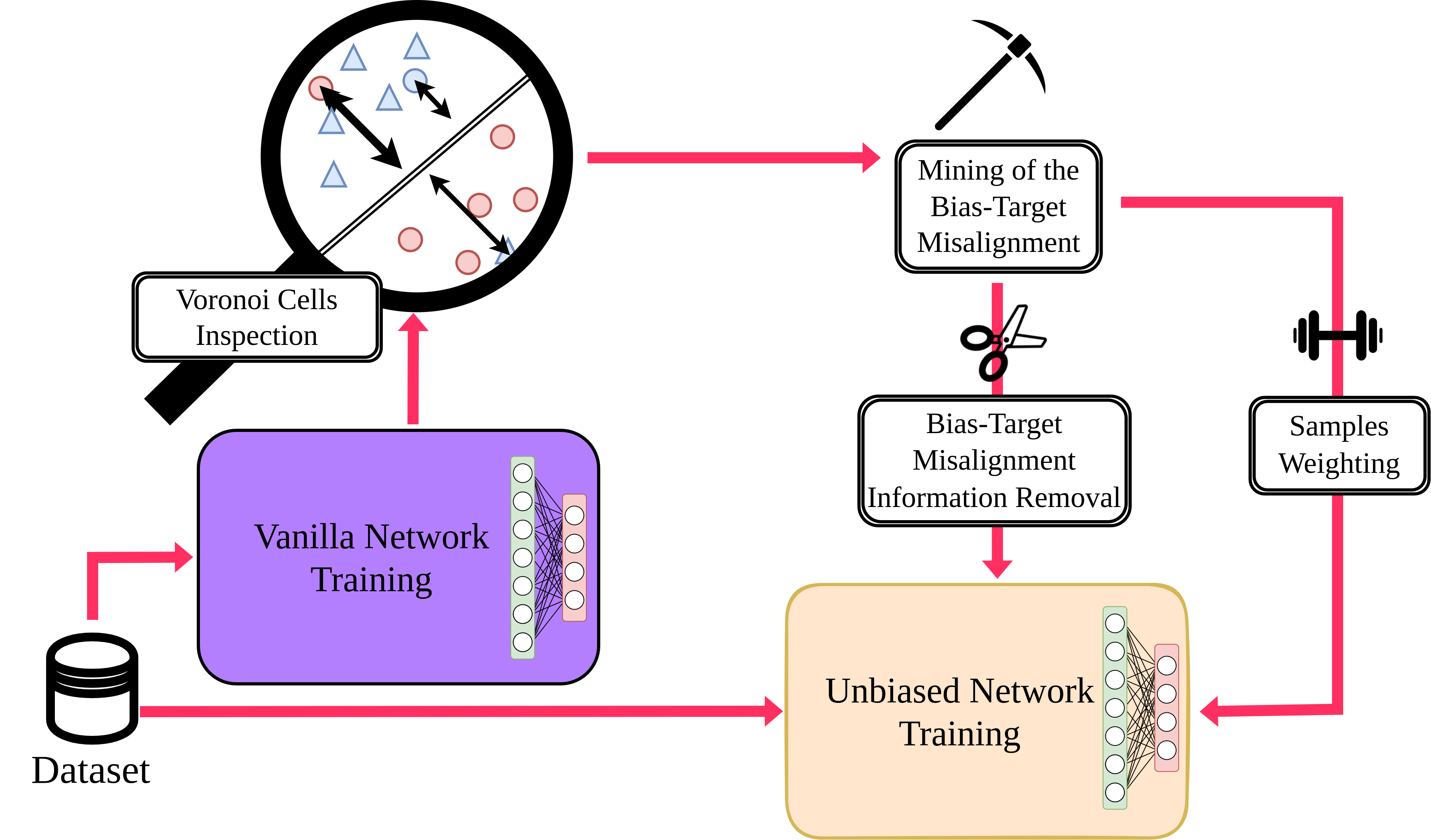}
\end{center}
   \caption{Our proposed approach to agnostically remove the bias.}
\label{fig:teaser}
\end{figure}

\noindent Many current debiasing approaches rely on prior information about the bias, such as the existence of an auxiliary label indicating some side information, the presence of bias(es) or their quality \cite{EnD,FairKL, Rebias,rubi,HEX,LearnedMixin}. However, obtaining these labels or information about the nature of the bias can be either very expensive (due to annotation costs) or very noisy: this is what motivates the development of bias-agnostic approaches. Recent works have shown that bias features are learned ``early''~\cite{LfF, DFA}: there are bias-target \emph{aligned} samples, for which the bias is learned, and the performance on the train set increases, and some \emph{disaligned} ones, for which the prediction is wrong. Since bias-agnostic approaches delve into the biased information from the training set, it is common to amplify the first features learned using Generalized Cross-Entropy (GCE)~\cite{gce} and then discourage their learning in an ``unbiased'' model. However, there is no guarantee that the very first features learned are the biased ones: detecting them agnostically and effectively remains an open question.

\noindent In this work, we propose a method that identifies the best time to extract bias-target alignment information by observing the relative distance of misclassified samples to the nearest Voronoi hyperplane of the correct target class. We use this information to train an unbiased model, from which we give higher weight to bias-misaligned samples, and remove the bias-alignment information from the bottleneck layer (Fig.~\ref{fig:teaser}).\\
At a glance, our contributions are the following:
\begin{itemize}[noitemsep,nolistsep]
	\item we propose a bias-agnostic approach which indicates, during the training of a vanilla model, when to extract bias-target alignment information: more precisely, we observe the distance of misclassified samples to the closest Voronoi hyperplane of the correct target class (Sec.~\ref{subsec:bias_inf});
	\item we use the bias misalignment information to weight the loss contribution of every single sample: this will favor the learning of misclassified samples in the vanilla setup (Sec.~\ref{sec:lfr});
	\item we also propose an approach to eliminate bias misalignment information: specifically, we minimize the bias alignment information which is extractable at the bottleneck of the DNN, conditioned to the bias target alignment (Sec.~\ref{sec:mir});
	\item we study the behavior on several datasets typically used for debiasing and compare both supervised and bias-agnostic approaches: although our proposed technique is bias-agnostic, its performance is comparable to supervised approaches (Sec.~\ref{sec:comparison}).
\end{itemize}
\section{Related works}
\label{sed:sota}
Avoiding algorithmic bias plays an important role in artificial intelligence (AI) ethics, and the area of research which focuses on this is \emph{fairness}. DNN predictions are ``unfair'' if they are based on specific sets of features and classes that will unfairly impact certain groups, according to some ethical principles~\cite{fairness_metrics,dwork_fairness_2012,fairlearn, InfoFair}. Some fairness metrics measure this type of inequality: Demographic Parity, Equalized Odds, or Equal Opportunity are some representative examples~\cite{EO, fairness_metrics, fairness_metrics2}. Fairness and debiasing are intrinsically related, but while the former explicitly highlights a measure of fairness, the latter maximizes the performance on an ``unbiased'' dataset, without necessarily declaring a fairness metric. Below, we will review debiasing approaches, which can be divided into \emph{supervised} methods (where we have access to a ``bias label''), and \emph{unsupervised} methods. 

\subsection{Supervised methods}
Supervised debiasing methods are divided into three categories: pre-processing methods, which modify the dataset prior to classification; in-processing methods, which modify the learning process of the model; and post-processing methods, which directly modify the output of the DNN.\\
\textbf{Preprocessing methods.} Among the most used preprocessing methods in the literature, driven data augmentation plays a prominent role. Generative Adversarial Networks (GANs) are widely used to generate realistic images: StyleGANs~\cite{StarGAN} is indeed one of the mostly used GANs in this context. 
For example, Kang~\etal~\cite{handwritting} used it to generate handwritten text in specific styles. 
In image classification, Geirhos~\etal~\cite{geirhos_texture} used style transfer to augment ImageNet with texture-bias-conflicting elements to create a more texture-balanced dataset.\\
\textbf{Postprocessing methods.} These methods have the advantage of neither re-training models nor requiring additional data for the training. 
With their Reject Option Classification, for example, Kamiran~\etal~\cite{ROC} proposed to take the samples classified with the most uncertainty (outside a predefined confidence margin) and to change their class to a lower Disparate Impact. 
In this same context, Equalized Odds Postprocessing proposed by Hardt~\etal~\cite{EO} maximizes the Equalized Odds metric. 
Despite the potential advantages of these approaches, a major drawback lies in the low degrees of freedom for the corrections (since they can only access post-classification information), which limits their practical effectiveness.\\
\textbf{In-processing: debiasing within training.} Most of the debiasing methods in the literature work directly on the model, learning from a biased dataset. In general, unbiased elements are weighted more than biased elements. This simple yet effective approach is nowadays very popular in supervised setups~\cite{reweighing_supervised}. Other methods tackle supervised debiasing by adding regularization terms during the training of the deep model, which is the case of methods such as EnD~\cite{EnD} and FairKL~\cite{FairKL}. Another intuitive approach relies upon simply removing the biased features from each sample in the dataset and performing the so-called \emph{fairness by blindness}. However, the phenomenon known as \emph{encoding redundancy}~\cite{EO} states that information is very rarely encoded only once in the data~\cite{fairness_mining}, so removing a single value or label is probably not sufficient to remove the effect of the bias on classification. 

\subsection{Unsupervised methods}
Some recent methods do not rely on bias labels because they can be difficult to obtain on real-life datasets and we will refer to them as ``unsupervised'' or ``bias-agnostic''. All of these approaches follow a general scheme, which is typically divided into two phases: \emph{bias inference}, where a first model, often called ``bias capturing'', aims to capture biases in the data; and \emph{bias mitigation}, where a second model is trained to avoid the biases captured by the first model. These approaches rely on prior knowledge, which may be more or less specific to the target task.\\
\textbf{Bias is in the texture.} Some works focus on the bias specifically present in texture, as it is prominent in image classification~\cite{geirhos_texture}. Rebias~\cite{Rebias} promoted learning with representations that are maximally different from using small receptive fields in convolutional layers. These are biased-by-design, toward learning specific textures. For the same texture debiasing task, HEX~\cite{HEX} proposed to use the gray-level co-occurrence matrix and to promote representation independent of colors.\\
\textbf{Bias generates imbalances between groups.} Some unsupervised approaches involve finding the bias groups that optimize some fairness metric and train the model to have representations orthogonal to those inferred by the biases. With DebiAN~\cite{Debian}, Li~\etal for example proposes a method that alternates between the training of bias-capturing and unbiased models, minimizing the Equal Opportunity fairness metric. In EIIL~\cite{EIIL}, Creager~\etal identified biases by finding the groups that maximize violation of an invariance principle measured by the objective function IRMv1~\cite{IRM}. Similarly, PGI~\cite{PGI} built upon EIIL by minimizing the KL-Divergence of the prediction over these groups.\\
\textbf{Bias is learned early.} Some recent methods are based on the assumption that bias features are easy to learn. These features can be extracted at a given point, at the beginning of the training. With LfF~\cite{LfF}, Nam~\etal proposed a loss reweighing method based on this assumption: they train a biased neural network and amplify its early stages prediction. In parallel, they train a debiased model weighting ``difficult samples''. Based on this, with DFA~\cite{DFA}, Lee~\etal performed data augmentation attempting to disentangle bias features from intrinsic features through latent representations of the bias-capturing and unbiased models. \\
The closest competing strategy to the one we propose is LfF. Differently from \cite{LfF}, our main assumption is not that the first features learned by the model are biased: we assume that the model, at some point, will adapt to the bias and that it is possible to identify this moment by examining the latent representation of the dataset. This particular point can occur at any time during the training, and so emphasizing the earlier choices of the model can prevent it from efficiently adapting to the bias. Moreover, unlike \cite{LfF}, we do not seek to extract the information from the bias, but its alignment with target classes, which allows our approach to easily scale to multi-biased setups. 
\section{Proposed Method}
\label{sed:method}
\begin{figure*}[hbt!]
\begin{center}
   \includegraphics[width=\linewidth]{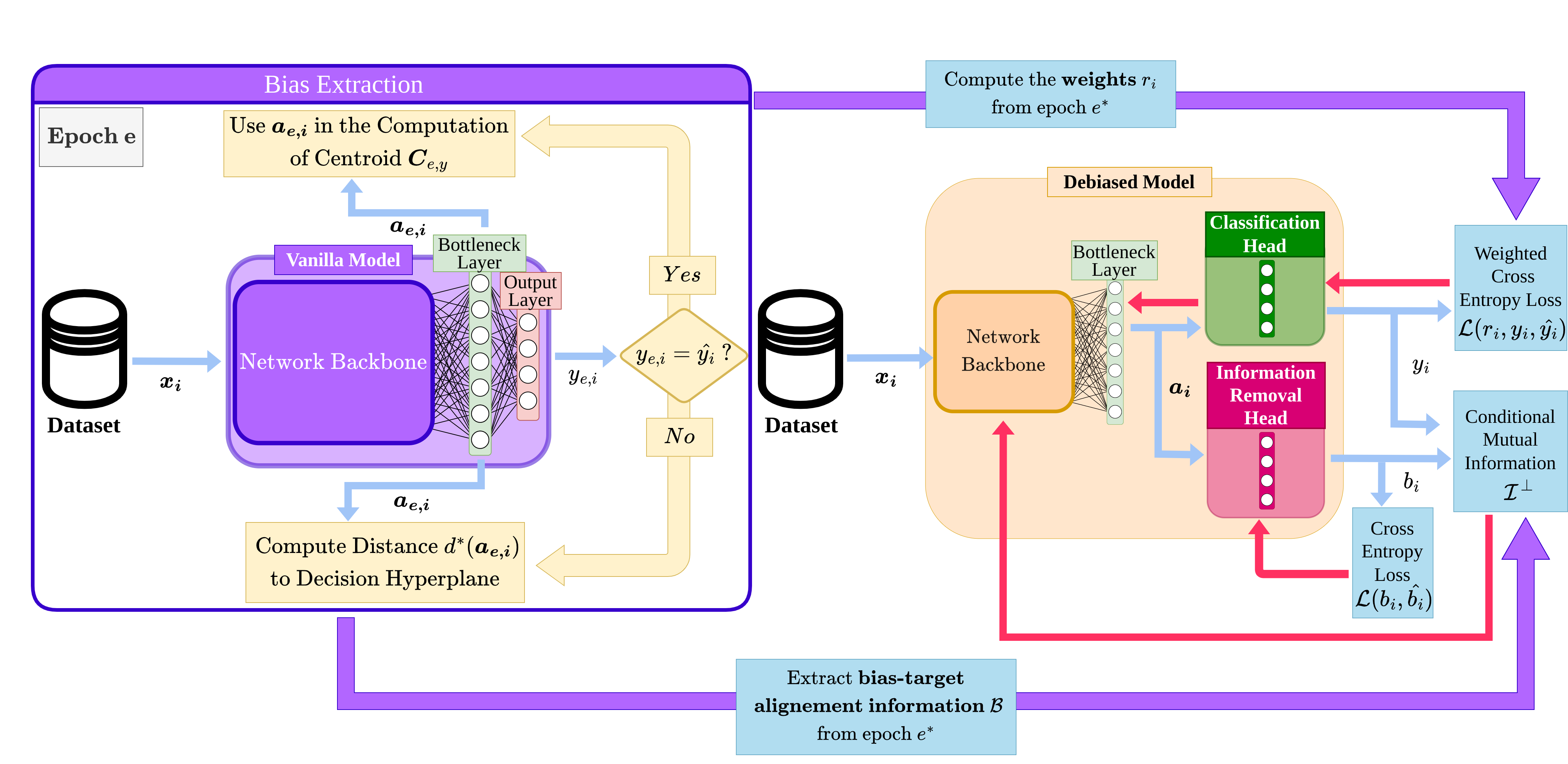}
\end{center}
   \caption{Overview of the proposed debiasing approach: the bias is first extracted (left) and then the unbiased model is trained (right). Blue arrows represent forward propagation while red arrows represent backpropagation.}
\label{fig:pipeline}
\end{figure*}
\subsection{Overview of the proposed method}
\label{sec:overview}
Let us consider a supervised learning setup, where we have a dataset $\mathcal{D}$ containing $n$ input samples $(\boldsymbol{x}_1,\!...,\!\boldsymbol{x}_n)\!\in\!\mathcal{X}$, each associated to a ground truth target label 
$(\hat{y}_1,\!...,\!\hat{y}_n)\!\in\!\mathcal{Y}$. A given deep neural network $\mathcal{M}$, trained for $e$ epochs, produces an output $y_{e,i}$ given some $\boldsymbol{x}_i$, and is typically trained to match $\hat{y_i}$ $\forall i$ through the minimization of a loss function $\mathcal{L}(y_{e,i}, \hat{y_i})$. Unfortunately, this learning process does not impose any prior on the specific subset of features that are extracted: these can lead the prediction over unseen data to be biased: we want to fight this effect.\\
Fig.~\ref{fig:pipeline} provides an overview of the proposed debiasing approach. First, the bias is inferred by the learning of a vanilla model: at the end of each epoch (or after a few iterations), the target class centroids and the decision hyperplanes are computed from the well-classified samples at the bottleneck layer (Sec.~\ref{sec:bottleneck}), and the distance of the misclassified samples to the Voronoi cell of the correct class is computed to find the epoch $e^*$ when the bias-target alignment is maximally learned (Sec.~\ref{subsec:bias_inf}). Then, a debiasing process follows (Sec.~\ref{sec:debiasing}): from the distances gathered from the previous step, we assign each sample a weight, which will be used in the weighted cross-entropy loss (Sec.~\ref{sec:lfr}). In addition, at the bottleneck layer, we also minimize the information about bias misalignments: this favors the unbiasedness of the classification head (Sec.~\ref{sec:mir}). In the rest of this section, we will detail all the steps of our proposed technique.

\subsection{Bottleneck latent representation}
\label{sec:bottleneck}

The debiasing method we propose stems from the concept of latent representation: the output of each layer of a DNN consists of a representation of the input $\boldsymbol{x}_i$. Thus, the classification phase, which takes place just before the output of the model, consists in partitioning the feature space into each of the different classes. Therefore, the output of the \textit{bottleneck layer} (the output of the backbone) is the compressed representation of the input sample, which is often referred to as \textit{latent representation}. Therefore, each $\boldsymbol{x}_i$ has a vector of latent attributes $\boldsymbol{a}_{e,i} = (a_{e,i,1},...,a_{e,i,K})\in \mathbb{R}^{K}$ associated to a specific epoch (or iteration) $e$ for the model $\mathcal{M}$: this forms its latent bottleneck representation.\\
We define $\mathcal{D}_{e}^{\mathbin{\|}}$ the set of samples well classified by the model $\mathcal{M}$ at epoch $e$, and $\mathcal{D}_{e}^{\perp}$ the misclassified samples. For each $t$-th target class, it is possible to define a \emph{class centroid} $\boldsymbol{C}_{e,t}$ as the average of the bottleneck representations of each well-classified samples of the $t$-th target class:
\begin{equation}
    \boldsymbol{C}_{e,t} = \frac{1}{ \| \mathcal{D}_{e,t}^{\mathbin{\|}}  \|_0} \sum_{i\in \mathcal{D}_{e,t}^{\mathbin{\|}}}\boldsymbol{a}_{e,i},
    \label{eq:centroids}
\end{equation}
where $\mathcal{D}_{e,t}^{\mathbin{\|}}$ is the subset of correctly classified samples for the $t$-th class, and $\|\mathcal{D}\|_0$ counts the elements in $\mathcal{D}$. Such centroids are proxies for the representations of the correctly classified elements of the class by the model. Let us define \textit{decision hyperplane} $\boldsymbol{H}_{e,i,j}$ as the hyperplane equidistant from $\boldsymbol{C}_{e,i}$ and $\boldsymbol{C}_{e,j}$ in the bottleneck representation space of $\mathcal{M}$: $\boldsymbol{H}_{e,i,j}$ is a proxy of the Voronoi decision boundary.\\
\begin{figure}[hbt!]
\begin{center}
\includegraphics[width=\linewidth]{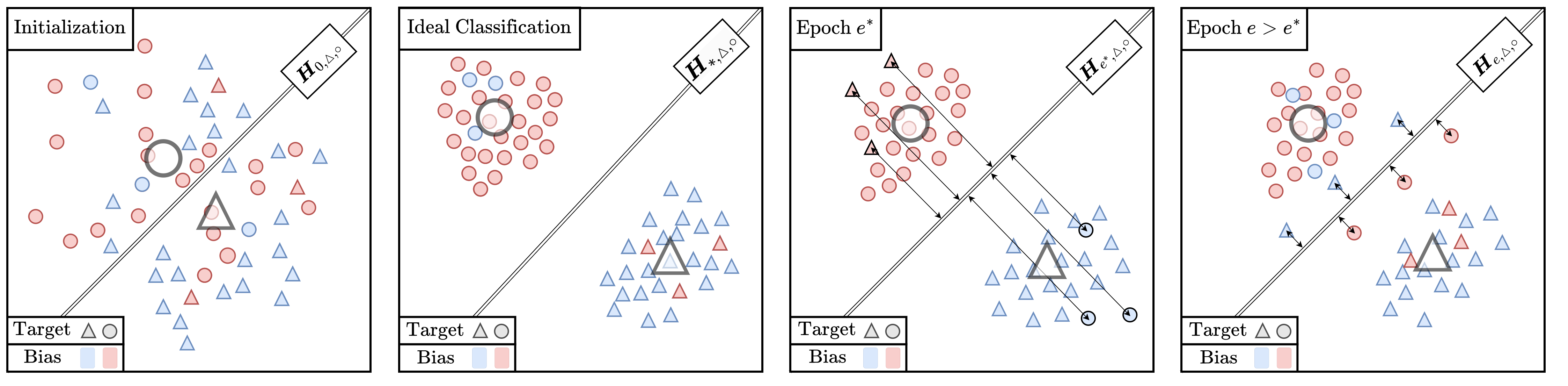}
\end{center}
   \caption{Representation of latent representations of a dataset at different learning stages. The target classes are shapes and the bias is color. Arrows represent the distance between specific elements and the Voronoi hyperplane.}
\label{fig:centroids}
\end{figure}

\noindent In Fig.~\ref{fig:centroids}, we have two target classes (the triangles and the squares) and the bias is pictured as the color (blue and red). The first image on the left shows the latent representation of the dataset by the model at its initialization: the samples are scattered randomly in the feature space, a first classification occurs, materialized by the decision hyperplane $\boldsymbol{H}_{0,\vartriangle,\circ}$. Ideally, as shown in the second picture for Fig.~\ref{fig:centroids}, a deep learning model should minimize the intra-class distance and maximize the inter-class one. However, in the presence of bias, another kind of attractor can emerge: the bias-conflicting elements will be attracted by the wrong class. The more the model is biased, the more the sample clusters that form will represent the biased classes more than the target class. For instance, in the third picture, the samples are clustered by color and not by shape: the red triangles have been attracted by the circle class that is correlated with the color red. If the model is, then, over-parametrized, the biased elements will be attracted as well towards the right centroid (fourth figure): although a set of bias-misaligned features are learned, the model still holds biased ones, which leads to a subpar generalization performance: our first goal will be, hence, to detect the $e^*$ moment of the learning where it is possible to extract the bias-target misalignment information.

\subsection{Bias alignment capture} 
\label{subsec:bias_inf}
To distinguish between bias-misaligned samples and bias-aligned ones, we assume that, after a few learning steps, the farther a misclassified sample is distant from its target Voronoi cell, the more it has been strongly pulled by an attractor. Such an attractor, since it is not its target class centroid, can be considered as resulting from some bias. Hence, when the average distance between the misclassified samples and their target Voronoi cell reaches its maximum, the model has learned bias features. To select the exact moment when to extract the bias-target alignment, we are looking for the epoch
\begin{equation}
    e^* = \argmax_{e} \frac{\sum_{i} d^*(\boldsymbol{a}_{e,i})}{\|\mathcal{D}_{e}^{\perp}\|_0} \\
\label{eq:bias_inf}
\end{equation}
where
\begin{equation} 
\label{eq:absdist}
d^*(\boldsymbol{a}_{e,i}) = 
\left\{\begin{array}{ll}
        0 &\text{if } y_{e,i} = \hat{y}_i\\
        \displaystyle\|\boldsymbol{a}_{e,i} - \boldsymbol{H}_{e,\boldsymbol{C}_{e,y_{e,i}},\boldsymbol{C}_{e,\hat{y}_i}}\|_2 &\text{if } y_{e,i} \neq \hat{y}_i,
\end{array}\right.
\end{equation}
where $\|\cdot\|_2$ indicates the $\ell_2$ norm. This scenario is visualized in Fig.~\ref{fig:centroids} (bottom-left). At this point, we can collect bias-target alignment information $(y_{e^*,1}\!=\!\hat{b}_{1},...,y_{e^*,n}\!=\!\hat{b}_{n})\!\in\!\mathcal{B}$. After this step, we can hereby identify the subset of bias-target misaligned samples $\mathcal{D}^\perp$, and the set of bias-target aligned ones $\mathcal{D}^{\mathbin{\|}}$. Given that vanilla learning strategies employ weight-decay, for which the distances tend to diminish when reaching the loss minimum, we propose to modify \eqref{eq:absdist} as
\begin{equation} 
\label{eq:reldist}
d^*(\boldsymbol{a}_{e,i}) = 
\left\{\begin{array}{ll}
        0 &\text{if } y_{e,i} = \hat{y}_i\\
        \displaystyle 2\cdot \frac{\|\boldsymbol{a}_{e,i} - \boldsymbol{H}_{e,\boldsymbol{C}_{y_{e,i}},\boldsymbol{C}_{e,\hat{y}_i}}\|_2}{\|\boldsymbol{C}_{y_{e,i}}\|_2 + \|\boldsymbol{C}_{e,\hat{y}_i}\|_2} &\text{if } y_{e,i} \neq \hat{y}_i,
\end{array}\right.
\end{equation}
where we scale the distance by the average norm of the two considered centroids.\\
We highlight that our hypothesis is similar but substantially different, from the one formulated in LfF~\cite{LfF}: here, we are free from the assumption bias features are learned earlier than all the more robust ones by the models, and even more from the assumption that they are the very first features learned by the models. We only state that at the point when the model learns these features, the misclassified elements will mainly be bias-conflicting: we can extract this information by monitoring the bias-target misalignments. 

\subsection{Debiasing the model}
\label{sec:debiasing}
Once having extracted the bias alignment labels for each sample, we can start training and debiasing our actual model. In this work, we propose unbiased models having the same architecture and learning parameters as the bias extractor one, although no explicit constraint forbids us to use different models. Our approach here consists of modifying the objective function optimized during the training of the model with two goals: increasing the weight of the bias-misaligned samples (as they contain relevant information on generalized n compared to other representatives of the same class) and moving these samples away from the bias centroid that tends to attract them. 

\subsubsection{Loss function reweighting}
\label{sec:lfr}
As shown by LfF~\cite{LfF} and other works like~\cite{reweighing_supervised}, reweighting the loss function to up-weigh the bias-conflicting elements is an efficient method to orient the training in a less biased direction as the correlation bias-target will be less emphasized in the loss. In this work, we assign the weight $r_i$ to the $i$-th sample according to
\begin{equation}
r_i = \left\{\begin{array}{cl}
    \displaystyle\frac{1}{\rho_{\hat{b}_i}} &\text{ if } x_i \in \mathcal{D}^{\mathbin{\|}}\\ ~\\
    \displaystyle\frac{1}{1-\rho_{\hat{b}_i}} & \text{ if } x_i \in \mathcal{D}^{\perp},
    \end{array}\right.  
\end{equation}
where we define 
\begin{equation}
    \rho_c = \frac{\left \|  \mathcal{D}_c^{\mathbin{\|}}\right \|_0} {\left \| \mathcal{D}_c \right \|_0}.
\end{equation}
In a nutshell, misaligned samples receive a weight that is proportionally inverse to their cardinality in the $c$-th class. This has the effect of strongly encouraging the learning of bias-misaligned samples, over bias-aligned ones. In the feature space, we can interpret the resulting reweighted loss $\mathcal{L}(y_i, \hat{y}_i, r_i)$  as an \emph{attractive force} on the bias-misaligned samples, that are being pulled toward their class centroids. 

\subsubsection{Bias alignment information removal}
\label{sec:mir}
Besides having a loss reweighting to favor misaligned sample learning, we can also discourage the model from learning any information related to bias alignment to the target. To estimate how much of this information is learned by the model, at the bottleneck we plug an auxiliary classification head that we call \emph{information removal head} (IRH). This head is trained to minimize a cross-entropy loss $\mathcal{L}(b_i, \hat{b}_i)$. Its performance is an important indicator for us, as reveals how much the latent space is similar (or different) from the vanilla bias-capturing model, when it was the most fitted to the bias (at epoch $e^*$), from the bias-target misalignment perspective. To be more accurate, this is estimated through the computation of the mutual information between $b_i$ and $\hat{b}_i$ under the bias misalignment condition
\begin{equation}
    \mathcal{I}^\perp =\sum_{j, k}p_{b,\hat{b}}^\perp(j,k)\log\left[\frac{p_{b,\hat{b}}^\perp(j,k)}{p_{b}^\perp(j) p_{\hat{b}}^\perp(k)}\right],
\label{eq:CI}
\end{equation}
where 
\begin{equation}
    p_{b,\hat{b}}^\perp(j,k) = \frac{\sum_i b_i \cdot \delta_{\argmax(b_i),j} \cdot \delta_{\hat{b}_i,k}}{\|\mathcal{D}^\perp\|_0}
\end{equation}
is the joint probability between $b$ and $\hat{b}$ calculated on the bias-target misaligned samples, $\delta_{i,j}$ is the Kronecker delta, and $p_{b}^\perp$, $p_{\hat{b}}^\perp$ are the two marginals. \eqref{eq:CI} is differentiable, as $b_i$ is the softmax-ed output of the IRH: hence, we are allowed to minimize this term, eventually scaled by a hyper-parameter $\lambda_{\mathcal{I}^\perp}$.\\
As also displayed in Fig.~\ref{fig:pipeline}, the mutual information does not contribute to the information removal head's update, but it is propagated directly back to the backbone. Minimizing the information over the bias-target misaligned samples can be seen as a \emph{repulsive force}, and should not be performed also on the bias-target aligned ones: indeed, we do not wish to destroy information related to the target class but to remove the link between bias misaligned samples and their attractor class. In the next section, we will test our approach and compare it with other state-of-the-art methods.
\section{Experiments}
\label{sed:exp}
Every result here presented is averaged over three seeds as done in most of the literature, and every algorithm is implemented in Python, using PyTorch~1.13, and trained on GPUs Nvidia GeForce RTX3090~Ti equipped with 24GB RAM. As we compare our method to the other unsupervised state-of-the-art methods, the best-unsupervised accuracies are systematically in bold and the second best are underlined. Besides, we also highlight in red the best method overall to see how our method compares even to the supervised ones.
The source code is available at \url{https://github.com/renahon/mining_bias_target_alignment_from_voronoi_cells/}.

\subsection{Datasets}
\label{sed:datasets}
\noindent Here below we describe at a glance the various datasets employed for the quantitative evaluation of our method.\\

\noindent\textbf{Biased MNIST.} The first dataset we are using is Biased MNIST, which was first introduced by Bahng~\etal~\cite{Rebias}. The 60k samples of this dataset consist of a colored version of the famous handwritten digits dataset MNIST with some correlation $\rho$ between the color and the digits. To build it, first one specific color gets assigned to each of the ten digits; then each of the samples gets its background color. We test four levels of color-digit correlation $\rho$: 0.99, 0.995, 0.997, and 0.999. The effect of the bias (namely, the background color) is evaluated by testing the model on a completely unbiased dataset, with $\rho=0.1$.  \\
\begin{figure}[tbh!]
\begin{center}
   \includegraphics[width=0.77\linewidth,trim={0 1cm 0 0},clip]{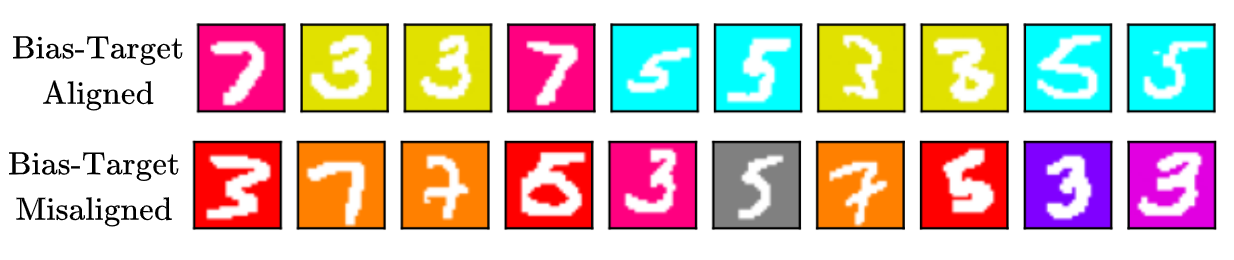}
\end{center}
\caption{Example of bias-aligned (top row) and bias-misaligned (bottom row) samples (of target values 3, 5, and 7) for the Biased-MNIST dataset \cite{Rebias}}
\label{bias_bmnist}
\end{figure}

\noindent In fig.\ref{bias_bmnist}, we can see an example of samples from this dataset that are either bias-aligned (top-row) or bias-misaligned (bottom-row). We can see on the top aligned row that each iteration of the same digit has the same color and that in the bottom misaligned one, the same digit has never that color. For example, all threes are yellow in the bias-aligned row whereas none of them is in the bottom row.\\
\noindent \textbf{Multi-Color MNIST.} Building on top of Biased MNIST, Li~\etal proposed in \cite{Debian} a bi-colored version to better benchmark the performance of current models on multiple biases at once. Here, the left and the right side of the background have two different colors, with a correlation to the target $\rho_L$ and $\rho_R$. We follow their proposed setup, with $(\rho_L,\rho_R) = (0.99,0.95)$.\\
\textbf{CelebA.} CelebA~\cite{CelebA} is a real-world dataset commonly used to test debiasing performance. It is a face classification dataset provided with 40 attributes for each of the 203k image samples. The task we solve here is to classify ``blond'' or ``not blond'' hair, with the main bias lying on the gender, as the dataset presents a natural bias for ''females'' to have the ''blond'' attribute.\\
\textbf{9-class ImageNet.} The 9-class ImageNet dataset was proposed by \cite{Rebias}, consisting of the extraction of a subset of 9 super-classes from ImageNet-1k, balanced to have each class correlated to a specific texture bias.\\
\textbf{ImageNet-A.} ImageNet-A was proposed by Hendrycks~\etal in \cite{ImageNet-A} as a subset of cropped images from ImageNet, purposely selected to be very hard to be classified by state-of-the-art CNNs. They were precisely selected among the subset misclassified by a cluster of ResNet50 models. Following~\cite{Rebias,biasCon,FairKL}, we use it to test our performance when training on 9-class ImageNet.

\subsection{Model architecture and training details}
For our experiments, we systematically used the same architecture for our vanilla model that allows us to measure the distances to decision hyperplanes and the model to debias. Therefore the architecture details are given below for each dataset stand for both models. Regarding the information removal head, we used SGD optimization with a learning rate of 0.1 for each model and dataset. Besides, we use $\lambda_{\mathcal{I}^\perp}=2$ for all our experiments.\\
Regarding the experiments on Biased MNIST, we used the same fully convolutional network used for Rebias~\cite{Rebias} and Irene~\cite{irene}, of four convolutional layers with $7\!\times\!7$ kernels, with a batch normalization after each of these layers. Regarding training, we followed the implementation used in~\cite{irene} of 80 epochs, with an initial learning rate of 0.1 decayed by 0.1 at epochs 40 and 60, and a weight decay of $10^{-4}$. For Multi-Color MNIST, we employ as architecture the same 3-layer MLP used in \cite{Debian}, trained for 500 epochs (until the model starts overfitting on the training set), using the same optimization strategy as in \cite{Debian}. For the experiments on CelebA and 9-class ImageNet, we used a pre-trained Resnet-18 and the same optimization strategy as in~\cite{LfF,biasCon}.

\subsection{Discussion}
\label{sec:comparison}
Here we compare our method to the current state-of-the-art, both supervised and unsupervised, on the multiple datasets presented in Sec.~\ref{sed:datasets}. In what follows, we divide the \emph{supervision level} of the method into three categories: bias-agnostic, bias-aware (using an extra ground-truth bias label), and ''bias tailored'' (BT) where the method does not rely on bias labels, but by construction, it captures specific biases (like the texture~\cite{Rebias,HEX}).\\
\begin{table}[htb!]
\centering
\caption{Results on Balanced Biased MNIST when training with different correlations color-digit $\rho$.}
\label{Bmnist}
\begin{tabular}{cccccc}
\toprule
\multirow{2}{*}{\bf Method} & \bf Bias & \multicolumn{4}{c}{\bf Test accuracy [\%] ($\uparrow$)}\\ 
       &  \bf agnostic    & \small{$\rho\!=\!0.999$}        & \small{$\rho\!=\!0.997$}      & \small{$\rho\!=\!0.995$}       & \small{$\rho\!=\!0.99$}       \\ 
       \midrule
Vanilla                          & \cmark & 11.2         & 40.5        & 72.4        & 88.4        \\ 
Rubi \cite{rubi}                    & \xmark & 13.7         & 90.4        & 43.0        & 93.6        \\
EnD \cite{EnD}                      & \xmark & 52.3         & 83.7        & 93.9        & 96.0        \\
BCon+BBal \cite{biasCon}      & \xmark & \hl{94.0}         & \hl{97.3}        & \hl{97.7}        & \hl{98.1}        \\ 
HEX \cite{HEX}                      & BT    & 10.8         & 16.6        & 19.7        & 24.7        \\
ReBias \cite{Rebias}                & BT    & 26.5         & 65.8        & 75.4        & 88.4        \\
LearnedMixin \cite{LearnedMixin}    & \cmark & 12.1         & 50.2        & 78.2        & 88.3        \\
LfF \cite{LfF}                      & \cmark & 15.3         & 63.7        & 90.3        & 95.1        \\
SoftCon \cite{biasCon}              & \cmark & \textbf{65.0}         & \underline{88.6}       & \underline{93.1}        & \underline{95.2}        \\
Ours                                & \cmark & \underline{58.7\!\small{$\pm$\!21.8}} & \textbf{92.7\!\small{$\pm$\!1.2}} & \textbf{95.5\!\small{$\pm$\!0.8}} & \textbf{97.7\!\small{$\pm$\!0.4}} \\ \bottomrule
\end{tabular}
\end{table}

\noindent \textbf{Results on Biased MNIST.} Our results on Biased MNIST, presented in Table~\ref{Bmnist}, show that our method achieves state-of-the-art performance, even when compared with supervised ones, for not extreme values of $\rho$: the only method that yields better results on the three lower correlations levels is the use of the associated BiasContrastive and BiasBalanced losses~\cite{biasCon}. On the unsupervised field, we get better accuracies than our competitors except for the highest correlation level. We can observe in this case a very high standard deviation because of the high stochastic noise of the gradient: the very few bias-target misaligned (60 in total, constituting the 0.1\% of the train set) searches for the perfect moment to mark. We hypothesize these difficulties are caused by the large gradients, which make the bias-target information extraction noisy: we tried to perform the bias extraction for this specific setup having a smaller learning rate ($0.01$), and the performance improved to 72.6\% $\pm$ 11.6. Tuning properly the learning rate in extreme scenarios is a key element towards a successful bias-target alignment information extraction.\\
\begin{table*}[htb!]
\centering
\caption{Test accuracy on four subsets of Multi-Color MNIST. The ``Unbiased'' one is the average of the four.}
\label{multimnist}
\begin{tabular}{cccccccc}
\toprule
\multirow{2}{*}{\bf Method} & \bf Bias & \multicolumn{5}{c}{\bf Test accuracy [\%] ($\uparrow$)}  \\
       & \bf agnostic   & \small{$A_{\text{left}}$ / $A_{\text{right}}$}  & \small{$A_{\text{left}}$ / $C_{\text{right}}$}  & \small{$C_{\text{left}}$ / $A_{\text{right}}$}          & \small{$C_{\text{left}}$  / $C_{\text{right}}$}      & \small{Unbiased} \\\midrule 
Vanilla                 & \cmark                 & 100.0        & 97.1        & 27.5        & 5.2         & 57.4                      \\
LfF \cite{LfF}                    & \cmark        & 99.6        & 4.7         & 98.6        & 5.1         & 52.0                      \\
EIIL \cite{EIIL}                   & \cmark       & 100.0        & 97.2        & 70.8        & 10.9        & 69.7                      \\
PGI  \cite{PGI}                   & \cmark                 & 98.6                    & 82.6        & 26.6        & 9.5         & 54.3                      \\
DebiAN  \cite{Debian}                & \cmark          & 100.0        & 95.6        & 76.5        & 16.0        & \underline{72.0}                      \\
Ours                    & \cmark                 & 100 \small{$\pm$\!0.0} & 90.9 \small{$\pm$\!3.5} & 77.5 \small{$\pm$\!2.8} & 24.1 \small{$\pm$\!1.8} & \hl{\textbf{73.1 \small{$\pm$\!0.9}}}               \\ \bottomrule
\end{tabular}%
\end{table*}

\noindent\textbf{Results on Multi-Color MNIST.} The Multi-Color MNIST dataset~\cite{Debian} helps us to test the performance of our method on multiple biases at the same time. Considering that there is a distinct correlation between the digits and the left color and them and the right background color, four accuracies are measured, regarding whether each background color is bias-aligned ($A_{\text{left}}$ for the left background color) or bias-conflicting ($C_{\text{right}}$ for the right background one). The most complex setup is $C_{\text{left}}-C_{\text{right}}$: it is below random guess for the vanilla model, but also three out of five tested debiasing methods. The average of these four metrics constitutes ``unbiased accuracy''. In contrast, every method reaches 100\% accuracy or close on the $A_{\text{left}}-A_{\text{right}}$ configuration. Even in this case, our method achieves state-of-the-art results for this dataset. More specifically, we record the best unbiased accuracy, and we improve the best score on the double-conflicting setup by the $+8\%$. For instance, LfF \cite{LfF}, whose main assumption is close to ours, while emphasizing the early choices of its bias-extracting model seems to perform very unevenly regarding the two biases (around 5\% accuracy when for $C_{\text{right}}$). This further strengthens our choice of not extracting biased information, but the bias-target alignment, and waiting for the best learning moment to extract such information.\\
\begin{table}[htb!]
\centering
\caption{Results on CelebA, targeting the attribute ``blond'', with a bias towards gender.}
\label{celebA}
\begin{tabular}{ccc c}
\toprule
\multirow{2}{*}{\bf Method} &  \multirow{2}{*}{\bf Bias agnostic} & \multicolumn{2}{c}{\begin{tabular}[c]{@{}c@{}}\bf Test accuracy [\%] ($\uparrow$)\end{tabular}} \\       
                &   & \small{Unbiased}     & \small{Bias-Conflicting} \\ \midrule
Vanilla                   & \cmark & 79.0  & 59.0     \\ 
EnD \cite{EnD}            & \xmark & 86.9  & 76.4       \\
LNL \cite{LNL}            & \xmark & 80.1   & 61.2      \\
DI \cite{DI}             & \xmark & 90.9  & 86.3       \\
BCon+BBal \cite{biasCon} & \xmark & \hl{91.4} & \hl{87.2}      \\
Group DRO \cite{groupdro}& \cmark & \underline{85.4}  & \underline{83.4}      \\
LfF \cite{LfF}            & \cmark & 84.2   & 81.2 \\ 
Ours            & \cmark & \textbf{90.2\!\small{$\pm$\!1.1}}  & \textbf{84.5\!\small{$\pm$\!2.0}}      \\ \bottomrule
\end{tabular}
\end{table}

\noindent\textbf{Results on CelebA.} On the CelebA dataset, two test setups are employed: the ''unbiased'', where the average of the scores obtained on each of the target-bias combinations (here blond-male, blond-female, not blond-male and not blond-female) is considered, and the ''bias-conflicting'' one, where the two bias-aligned combinations are not considered. On both the metrics, our approach ranks the best unsupervised (4.8\% more on the unbiased metric), and the third overall.\\
\begin{table}[htb!]
\centering
\caption{Test accuracy on 9-class ImageNet and ImageNet-A.}
\label{ImageNet}
\begin{tabular}{ccccc}
\toprule
\multicolumn{2}{c}{\multirow{2}{*}{\bf Method}}     & {\multirow{2}{*}{\bf Bias agnostic}} & \multicolumn{2}{c}{\bf Test accuracy [\%] ($\uparrow$)} \\ 
\multicolumn{2}{c}{}                 &       & \small{9-class ImageNet} & \small{ImageNet-A}  \\ \midrule
\multicolumn{2}{c}{Vanilla}         & \cmark & 94.0     & 30.5 \\
\multicolumn{2}{c}{ReBias \cite{Rebias}}          & \xmark & 94.0     & 30.5  \\
\multicolumn{2}{c}{StylImageNet \cite{geirhos_texture}} & BT    & 88.4     & 24.6 \\
\multicolumn{2}{c}{LearnedMixin \cite{LearnedMixin}}    & BT    & 79.2      & 19.0  \\
\multicolumn{2}{c}{RUBi \cite{rubi}}           & BT    & 93.9      & 31.0 \\
\multicolumn{2}{c}{LfF \cite{LfF}}             & BT    & 91.2      & 29.4 \\
\multicolumn{2}{c}{SoftCon \cite{biasCon}}         & BT    & 95.3     & 34.1 \\
\multicolumn{2}{c}{FairKL \cite{barbano}}     & BT                          & 95.1     & \hl{35.7}     \\
\multicolumn{2}{c}{Ours (BagNet \cite{bagnet})} & BT                          & \hl{96.4 \small{$\pm$\!0.0}}     & 34.5 \small{$\pm$\!3.4}    \\ 
\multicolumn{2}{c}{Ours}             & \cmark & \textbf{95.5 \small{$\pm$\!0.2}}     & \textbf{34.2 \small{$\pm$\!0.9}}
\\ \bottomrule
\end{tabular}%
\end{table}

\noindent\textbf{Results on 9-class ImageNet.} When working at debiasing 9-class ImageNet, all other state-of-the-art methods become bias-tailored: indeed, they use BagNet18~\cite{bagnet} as a bias-extracting model for its known tendency to fit texture because of its small receptive fields (as 9-class ImageNet and ImageNet-A are known to be very biased towards texture), instead of ResNet18 which is the model we are trying to de-bias. If we want to perform bias-agnostic debiasing, we shouldn't rely on that kind of bias-extracting model chosen to fit the bias type of the dataset. However, to compare our method to theirs on an equal footing we tested two configurations: 
\begin{itemize}[noitemsep,nolistsep]
    \item ours (+BagNet), where we extract the bias-conflicting samples from training BagNet18 and then proceeded to debiasing ResNet18;
    \item ours, where we extract the bias-conflicting samples directly from the ResNet18, which makes us the only bias-agnostic method tested on this dataset.
\end{itemize}
We obtain comparable results to the state-of-the-art on ImageNet-A (respectively 1.2 and 1.5\% below the best-performing method in FairKL~\cite{FairKL}) and the two best results overall on 9-class ImageNet. Interestingly, we get state-of-the-art results with our method, when extracting information directly from ResNet18.\\

\begin{table}[htb!]
\centering
\caption{Ablation study on Biased~MNIST with $\rho=0.99$.
}
\label{Ablation1}
\begin{tabular}{ccccc}
\toprule
 \begin{tabular}[c]{@{}c@{}}\bf Weighted $\mathcal{L}$ \end{tabular} &
  \begin{tabular}[c]{@{}c@{}} \bf Information removal head  \end{tabular} &
  \begin{tabular}[c]{@{}c@{}} \bf Misaligned only \end{tabular} &
  \multicolumn{1}{c}{\begin{tabular}[c]{@{}c@{}} \bf  Test accuracy [\%]($\uparrow$)\end{tabular}} \\ 
  \midrule
&  &  & 88.4 \small{$\pm$\!0.5}          \\
\cmark &  & & 95.5 \small{$\pm$\!0.5}          \\
\cmark & \cmark & & 97.2 \small{$\pm$\!0.4}         \\
\cmark & \cmark & \cmark & \textbf{97.7 \small{$\pm$\!0.4}} \\
\bottomrule
\end{tabular}%
\end{table}

\noindent\textbf{Ablation study.} We tested the effect of the different modules of our method on Biased MNIST, training with $\rho = 0.99$. The results in Table~\ref{Ablation1} show that the use of the reweighted loss function $\mathcal{L}$ yields an average increase in accuracy of 7.1\% and that the use of the information removal head (IRH) further increases our score of 1.7\% more. Finally, employing a conditional mutual information term (on the bias-target misaligned elements only) in place of total information removal provides an extra gain in performance.\\
\begin{figure}[htb!]
\begin{center}
   \includegraphics[width=0.8\linewidth]{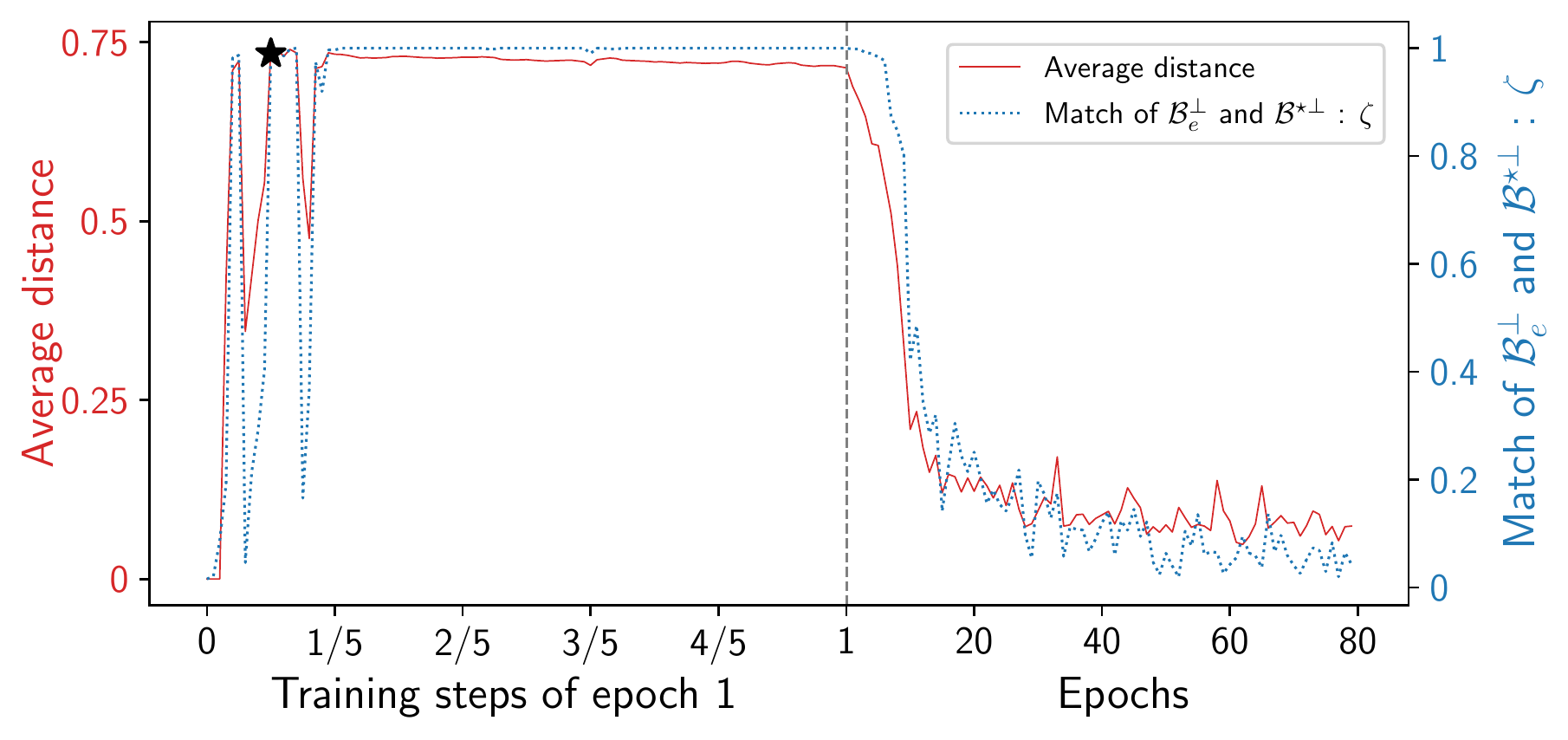}
\end{center}
\caption{Evolution of the relative distance to the target Voronoi cell for the misclassified elements (red curve) and bias alignment information match with the ground truth $\mathcal{B}^*$ (blue dashed line).}
\label{fig:dist}
\end{figure}

\noindent\textbf{Relevance of the Voronoi distance metric. } We also measured the average relative distances from the misclassified samples at each epoch, comparing it to the match of the vector of inferred bias labels $\mathcal{B}_e = (\argmax(b_{e,i}))_i^n$ to the ideal ground truth of bias labels $\mathcal{B}^*$ (which is provided in Biased~MNIST). It's computed over the set of misclassified samples $D^{\perp}$ as the bias-misaligned elements are our work's main point of interest. The value represented by the blue dashed line represents therefore the bias alignment information match and is computed as: 
\begin{equation}
\zeta = \frac{\sum_{i|x_i\in \mathcal{D}^\perp} \displaystyle \mathbb{1}_{\argmax(b_{e,i})=b^\star_i}}{\displaystyle|\mathcal{D}^\perp|}
\end{equation}
As shown in Fig.~\ref{fig:dist}, we can see a peak after the first few iterations, as the model fits the color bias with even less one epoch of training. We have marked with $\star$ the peak over the average relative distances. As the training goes on, the average relative distance goes down as expected: the misclassified samples stay closer to the decision hyperplane. We observe that the relative average distance is a good proxy for knowing when to learn the optimal bias alignment $\mathcal{B}^*$, as the two curves show a similar trend.\\
\textbf{General discussion and limitations.} Through the conducted experiments, we have observed that our method establishes, in most of the considered setups, a new state-of-the-art for bias-agnostic approaches, and in some cases even outperforms supervised methods, such as in 9-class ImageNet and the double-biased Multi-Color MNIST. A limitation of the proposed approach appears when the correlation between bias and target is extremely high ($\rho\!=\!0.999$ in Biased~MNIST). Since it heavily relies on the extraction of these bias-conflicting samples, when the stochastic noise overwhelms the extraction of the bias misalignment information, evidently the proposed method will be sub-optimal. A possible solution to this problem relies upon the use of a ``sufficiently small'' learning rate. Finally, our method strongly depends on the existence of these bias-conflicting elements: in a fully-biased dataset, where the alignment bias-target $\rho\!=\!1$, since we have no information to extract from the training set, our approach is expected to fail.

\section{Conclusion}
\label{sed:conclusion}
In this paper, we presented an unsupervised, bias-agnostic debiasing approach, whose performance is typically in the same range as state-of-the-art supervised methods. We proposed a new bias-target alignment extraction method based on the distance between the misclassified samples and the closest Voronoi hyperplane separating them from their target class. Based on this distilled information, we proposed a debiasing method consisting of two synergetic elements. The first consists of a reweighted cross-entropy loss, where the weights of the samples reflect the bias-target (mis)alignment. The second is a bias-target misalignment information removal term, acting as a regularizer for the latent space.
We tested our method on several debiasing benchmarks, recording a new state-of-the-art for unsupervised debiasing in most of the considered scenarios, although no specific hyper-parameters tuning has been performed. In extreme cases, where the bias-target alignment is extremely high, we have observed that the proper choice of the vanilla model's learning setup is crucial for the success of the proposed approach, and its exploration is left as future work.

\subsection*{Acknowledgments}
This work was granted access to the HPC resources of IDRIS under the allocation 20XX-AD011014080 made by GENCI.
{\small
\bibliographystyle{plain}
\bibliography{main}
}

\end{document}